\documentclass{article}
\usepackage{iclr2027_conference,times}

\usepackage{amsmath,amsfonts,bm}

\def\eqref#1{equation~\ref{#1}}

\def\1{\bm{1}}

\DeclareMathAlphabet{\mathsfit}{\encodingdefault}{\sfdefault}{m}{sl}
\SetMathAlphabet{\mathsfit}{bold}{\encodingdefault}{\sfdefault}{bx}{n}

\usepackage{hyperref}
\usepackage{url}
\usepackage{amsmath,amssymb}
\usepackage{graphicx}
\usepackage{booktabs}
\usepackage{multirow}
\usepackage{pifont}
\usepackage{xcolor}
\usepackage{enumitem}
\usepackage{float}

\title{ForensicZoom: Adaptive Visual Inspection with Multimodal LLMs for Industrial-Grade Face Forgery Detection}

\author{%
    Hang Zhou$^{*}$ \quad Yiming Tang$^{*}$ \quad Kun Yu \quad Qian Zhu \quad Minghao Li \quad Weigao Wen \\[0.3em]
    \hfill \textbf{Alibaba} \hfill\null \\[0.3em]
    \texttt{\{mathison.zh, fuming.tym, weigao.wwg\}@alibaba-inc.com} \\[0.5em]
    {\small $^{*}$Equal contribution.}
}

\iclrfinalcopy
\begin{document}

\maketitle

\begin{abstract}
Reliable face forgery detection is critical to the security of online identity verification systems, where missed attacks compromise security and excessive false positives disrupt legitimate users. Specialized forensic detectors achieve strong detection performance but provide limited interpretability, while multimodal large language models (MLLMs) offer strong semantic understanding and interpretable reasoning yet remain substantially weaker for face forgery detection. We argue that a key limitation lies in how visual evidence is acquired: subtle forensic artifacts may be poorly represented at standard resolution, while uniformly processing all cases at higher resolution is computationally inefficient. We therefore introduce \textbf{ForensicZoom}, an industrial-grade MLLM framework for \emph{adaptive visual inspection}. ForensicZoom first equips a general-purpose MLLM with forensic-aware visual representations and aligns the language model with these features. Its central mechanism, \texttt{NEED\_ZOOM}, enables the model to autonomously request magnified views of suspicious regions when the initial evidence is insufficient, turning fixed-pass classification into adaptive multi-round forensic reasoning. The zoom behavior is learned through reward shaping that balances detection accuracy with unnecessary visual inspection, concentrating additional computation on difficult cases. A final attribution optimization stage improves natural-language forensic reports while preserving detection performance. On large-scale industrial identity verification data, ForensicZoom achieves over $97\%$ TPR at $0.1\%$ FPR, substantially outperforming both specialized detectors and existing MLLM-based methods while producing actionable forensic attributions. These results demonstrate that ForensicZoom can provide an effective path toward accurate, interpretable, and scalable MLLM-based face forgery detection. Our scripts and model parameters will be released upon acceptance.
\end{abstract}
\section{Introduction}
\label{sec:intro}

As generative AI becomes increasingly capable and widely deployed, concerns over the safety and misuse of AI-generated content continue to grow~\cite{goodfellow2014generativeadversarialnetworks,ho2020denoisingdiffusionprobabilisticmodels,rombach2022highresolutionimagesynthesislatent}. Face forgery is a particularly consequential instance of this problem: malicious actors can manipulate or synthesize facial imagery to impersonate legitimate users and bypass online identity verification systems, creating substantial risks for financial services, e-government platforms, and other security-sensitive applications~\cite{sumsub2024fraud,jia2024facevlm,deepfakeeval2024}. Reliable face forgery detection is therefore essential, especially in industrial settings where missed attacks directly compromise security while excessive false positives disrupt legitimate users. Beyond accuracy, interpretability is important for supporting human review of ambiguous cases with clear forensic evidence. This challenge is becoming increasingly difficult as modern generative models produce highly realistic faces whose forensic traces may be limited to subtle local artifacts, further obscured by compression, heterogeneous capture conditions, and rapidly evolving generation pipelines~\cite{deepfakeeval2024,jiang2020deeperforensics}.

To address this problem, existing approaches primarily rely on specialized forensic models or multimodal large language models (MLLMs). Specialized CNN- and ViT-based detectors, including XceptionNet~\cite{chollet2017xception}, EfficientNet~\cite{tan2019efficientnet}, SBI~\cite{shiohara2022sbi}, and LAA-Net~\cite{laanet2024}, achieve strong detection performance but offer limited interpretability, typically producing predictions without explicit forensic reasoning or natural-language attribution. In contrast, MLLMs can jointly process visual and contextual information while providing interpretable explanations, but general-purpose MLLMs remain substantially weaker than specialized detectors for face forgery detection~\cite{jia2024facevlm,idaware2025}. Recent methods improve MLLM-based forensics through task-specific fine-tuning, expert-assisted prediction, artifact-aware supervision, reinforcement learning, and external forensic tools~\cite{huang2024ffaa,fakevlm2025,forgerygpt2024,mare2026,evolvereason2026,forgeryvcr2026}. Nevertheless, current MLLM-based methods still struggle to capture the fine-grained forensic evidence required for reliable detection and remain far from the stringent accuracy and efficiency requirements of industrial deployment.

Our experience with large-scale identity verification suggests that a central limitation lies in how visual evidence is acquired and allocated during inference. Human vision is known to follow a coarse-to-fine process, forming an initial global perception before allocating focused attention to local details when necessary~\cite{NAVON1977353,TREISMAN198097,GERLACH2020104131}. Recent MLLM research has begun to adopt a similar principle through guided visual search and active perception, allowing models to selectively inspect task-relevant regions rather than processing all visual information uniformly~\cite{wu2023vguidedvisualsearch,zhu2026activeo3empoweringmllmsactive}. Face forgery detection naturally exhibits this structure: many cases can be resolved from a global observation, while difficult cases may depend on subtle local artifacts that fall below the effective resolution of standard visual encoding. Applying high-resolution analysis uniformly is computationally inefficient at production scale, whereas fixed-resolution single-pass inference may overlook critical forensic evidence. These observations motivate an \emph{adaptive visual inspection} paradigm, where an MLLM selectively acquires higher-resolution evidence from suspicious regions when needed.

A key limitation of existing face forgery benchmarks is that they often evaluate models on isolated media inputs under distributions that do not fully reflect real-world deployment conditions~\cite{Chandra_2026_CVPR,Li_2026_CVPR}. Recent studies have increasingly emphasized multimodal evaluation, incorporating complementary visual, acoustic, depth, or infrared signals to better approximate practical face-security settings~\cite{Yu_2023_CVPR,shi2025shield}. In practice, however, behavioral information associated with the acquisition process---such as the time spent capturing an image, repeated verification attempts, and other interaction signals---can provide critical complementary evidence for distinguishing sophisticated attacks from legitimate users, particularly at the extremely low false-positive rates required in production. The absence of such signals creates a substantial gap between benchmark evaluation and industrial deployment: a pipeline designed and optimized solely for existing visual benchmarks may not directly translate to a practical verification system where visual and behavioral evidence must be jointly considered. To study face forgery detection under this more realistic setting, we therefore evaluate on a private multimodal dataset collected from an operational online identity verification system serving thousands of requests per day, which contains both visual inputs and associated behavioral information.

In this work, we introduce \textbf{ForensicZoom}, an industrial-grade MLLM framework for adaptive face forgery detection. We adopt an MLLM rather than a conventional vision-only detector because practical identity verification requires not only accurate predictions but also interpretable forensic evidence that can support subsequent human review \citep{rudin2019stop,guo2025rethinking,pmlr-v267-yu25d}. ForensicZoom first equips a general-purpose MLLM with forensic-aware visual representations and aligns the language model with these features. We further observe a substantial performance gap between global and locally magnified views for subtle forgery cases, indicating that critical forensic cues can be lost under fixed-resolution visual encoding \citep{zhao2021multi,liu2021spatial}. Its central mechanism, \texttt{NEED\_ZOOM}, therefore enables the model to autonomously request magnified views of suspicious regions when the initial evidence is insufficient, turning fixed-pass classification into adaptive multi-round forensic reasoning. The zoom behavior is learned through reward shaping that balances detection accuracy with unnecessary visual inspection, concentrating additional computation on difficult cases. A final attribution optimization stage improves natural-language forensic reports while preserving detection performance. On large-scale industrial identity verification data, ForensicZoom achieves over $97\%$ TPR at $0.1\%$ FPR while producing actionable forensic attributions, and has been deployed in a production pipeline processing millions of requests per day.

We summarize our contributions as follows:
\begin{itemize}
    \item We introduce \texttt{NEED\_ZOOM}, a learned mechanism that enables an MLLM to recognize insufficient forensic evidence and autonomously request higher-resolution inspection of suspicious regions.
    \item We develop \textbf{ForensicZoom}, a progressive training framework that builds forensic perception, aligns it with language reasoning, and enables adaptive multi-round inference with interpretable attribution.
    \item We validate ForensicZoom on large-scale industrial identity verification data, achieving over $97\%$ TPR at $0.1\%$ FPR, and demonstrate its effectiveness through production deployment at millions of requests per day.
\end{itemize}
\section{Related Work}
\label{sec:related}

\subsection{Face Forgery Detection}
\label{sec:related_ffd}

Face forgery refers to manipulated or generated facial imagery that alters a person's identity, appearance, or authenticity. Early detection methods exploit low-level forensic cues such as inconsistent head poses~\citep{yang2019headpose}, abnormal eye blinking~\citep{li2018blink}, and frequency artifacts~\citep{qian2020frequency}. Deep learning approaches learn forensic representations directly using architectures such as XceptionNet~\citep{chollet2017xception}, EfficientNet~\citep{tan2019efficientnet}, and Vision Transformers~\citep{coccomini2022vit4deepfake}. SBI~\citep{shiohara2022sbi} improves generalization through pseudo-forgery synthesis. LAA-Net~\citep{laanet2024} focuses on localized manipulation artifacts. FSFM~\citep{fsfm2025} learns self-supervised facial representations. Frequency-based methods explicitly model spectral inconsistencies~\citep{qian2020frequency,liu2021spatialphase}. More recently, MLLMs have been explored for interpretable forgery detection. DD-VQA~\citep{ddvqa2024} formulates detection as visual question answering, FFAA~\citep{huang2024ffaa} incorporates an external expert classifier, FakeVLM~\citep{fakevlm2025} fine-tunes on artifact annotations, and ForgeryGPT~\citep{forgerygpt2024} introduces identity-aware representations. MARE~\citep{mare2026} applies RLHF for attribution alignment. EvolveReason~\citep{evolvereason2026} uses GRPO for self-evolving forensic reasoning. \citet{sbirldeepfake2026} combines reinforcement learning with Self-Blended Images. ForgeryVCR~\citep{forgeryvcr2026} trains an MLLM to invoke external forensic tools.

\subsection{AI Content Safety}
\label{sec:related_safety}

The rapid development of generative models, from GANs~\citep{goodfellow2014generativeadversarialnetworks,karras2019stylebasedgeneratorarchitecturegenerative} to diffusion models~\citep{ho2020denoisingdiffusionprobabilisticmodels,rombach2022highresolutionimagesynthesislatent}, has substantially improved the realism and diversity of AI-generated visual content, creating growing challenges for content authenticity and safety. Early detection methods identify synthetic images through low-level statistical artifacts, including color inconsistencies~\citep{mccloskey2018detectinggangeneratedimageryusing} and generator-specific fingerprints~\citep{yu2019attributingfakeimagesgans}. Subsequent work investigates generalization across unseen generators~\citep{wang2020cnngeneratedimagessurprisinglyeasy,ojha2024universalfakeimagedetectors}, while GenImage~\citep{zhu2023genimagemillionscalebenchmarkdetecting} provides a large-scale benchmark for evaluating AI-generated image detection across diverse generative models. Diffusion-specific approaches further exploit reconstruction characteristics of diffusion models, such as DIRE~\citep{wang2023dirediffusiongeneratedimagedetection} and AEROBLADE~\citep{ricker2024aerobladetrainingfreedetectionlatent}. Beyond post-hoc detection, watermarking methods such as Tree-Ring Watermarks~\citep{wen2023treeringwatermarksfingerprintsdiffusion} embed identifiable signals directly into generated images to facilitate provenance verification. Mechanistic interpretability modules have been utilized to analyze visual patterns associated with generative-model failures: including Language-Grounded Sparse Encoders for general content analysis~\citep{tang2026humanlikecontentanalysisgenerative}, and Matryoshka Transcoders for physical plausibility failure modes~\citep{tang2025doesmodelfailautomatic}.

\section{Method}
\label{sec:method}

\begin{figure*}[t]
\centering
\includegraphics[width=\textwidth]{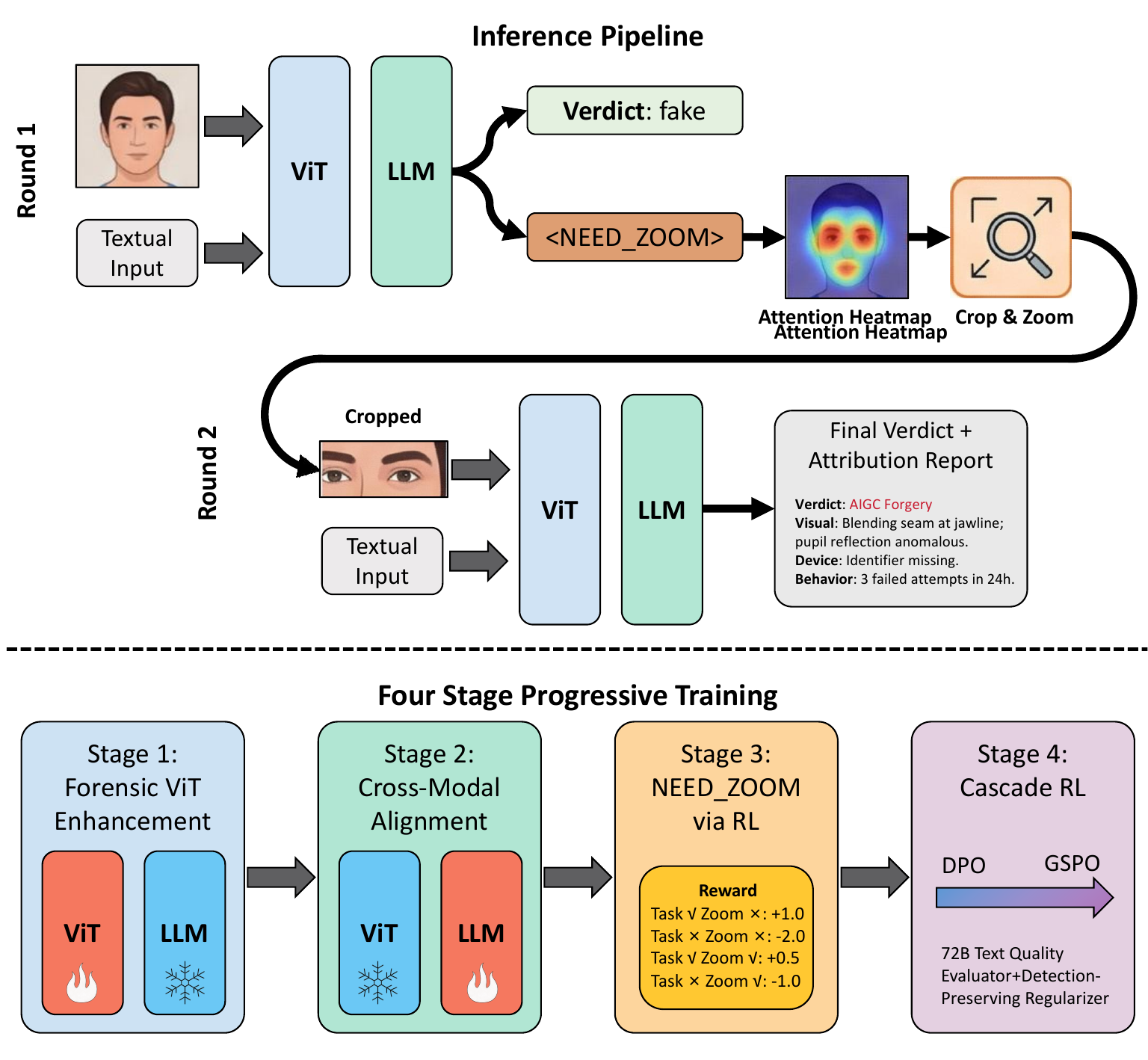}
\vspace{-5mm}
\caption{Overview of ForensicZoom. \textbf{Top:} Inference pipeline. In Round~1, the face image and text prompt are processed by the ViT (with Forensic Query Bank) and LLM. If the model outputs \texttt{<NEED\_ZOOM>}, the attention heatmap guides cropping and magnification of suspicious regions. In Round~2, the cropped patches are fed back through the same pipeline to produce the final verdict. \textbf{Bottom:} Four-stage progressive training. Stage~1 enhances the ViT with forensic perception. Stage~2 aligns the LLM with enhanced visual signals. Stage~3 trains the \texttt{<NEED\_ZOOM>} decision policy via PPO with asymmetric rewards. Stage~4 optimizes attribution quality through Cascade RL, transitioning from DPO to GSPO with a text quality evaluator and detection-preserving regularizer.}
\label{fig:pipeline}
\end{figure*}

In this section, we introduce the methodology of ForensicZoom (Figure~\ref{fig:pipeline}). Section~\ref{sec:stage1} first enhances the vision encoder with forensic perception, and Section~\ref{sec:stage2} aligns the language model with these forensic visual signals through supervised fine-tuning. Section~\ref{sec:stage3} then introduces the \texttt{NEED\_ZOOM} mechanism, which enables adaptive higher-resolution inspection. Finally, Section~\ref{sec:stage4} presents Cascade RL for improving forensic attribution while preserving detection performance.

\subsection{Forensic Vision Enhancement}
\label{sec:stage1}

We first adapt the vision encoder to capture the fine-grained, high-frequency cues that distinguish genuine faces from forgeries, and compress them into a compact set of forensic features. In specific, we freeze the LLM $\mathcal{L}$ and the projection layer $\mathcal{P}$, and unfreeze the last $K$ layers of the ViT encoder $\mathcal{V}$, approximately 25--30\% of total ViT layers. Let $\mathcal{V} = [\mathcal{V}^{(1)}, \ldots, \mathcal{V}^{(L)}]$ denote the $L$-layer ViT, where layers $1$ through $L-K$ remain frozen and layers $L-K+1$ through $L$ are trainable. Lower layers capture general visual features such as edges and textures that transfer well, while upper layers encode task-specific representations that require adaptation.

\textbf{Forensic Query Bank (FQB).}
We introduce a set of $M$ learnable query tokens $\mathbf{Q} = \{\mathbf{q}_1, \ldots, \mathbf{q}_M\} \in \mathbb{R}^{M \times d}$, where $d$ is the ViT hidden dimension, trained to extract forensic-relevant information from the ViT output through a cross-attention mechanism.

Let $\mathbf{Z} = \mathcal{V}(\mathbf{I}) \in \mathbb{R}^{N \times d}$ denote the $N$ visual tokens output by the ViT. The FQB applies multi-head cross-attention:
\begin{equation}
\label{eq:fqb}
\mathbf{F} = \text{CrossAttn}(\mathbf{Q}, \mathbf{Z}, \mathbf{Z}) = \text{softmax}\!\left(\frac{\mathbf{Q} \mathbf{W}_Q (\mathbf{Z} \mathbf{W}_K)^\top}{\sqrt{d_k}}\right) \mathbf{Z} \mathbf{W}_V,
\end{equation}
where $\mathbf{W}_Q, \mathbf{W}_K, \mathbf{W}_V \in \mathbb{R}^{d \times d_k}$ are learnable projection matrices and $\mathbf{F} \in \mathbb{R}^{M \times d}$ are the resulting \emph{forensic tokens}. These tokens ``query'' and disentangle the high-frequency forensic features from the entangled ViT representations, where semantic and forensic information are intertwined.

The forensic tokens are concatenated with the original visual tokens before being fed into the LLM:
\begin{equation}
\label{eq:concat}
\mathbf{H} = [\mathcal{P}(\mathbf{Z}); \, \mathcal{P}_f(\mathbf{F})] \in \mathbb{R}^{(N+M) \times d_l},
\end{equation}
where $\mathcal{P}$ and $\mathcal{P}_f$ are MLP projection layers mapping visual and forensic tokens respectively into the LLM embedding space of dimension $d_l$.

\textbf{Attention Entropy Guidance.}
To encourage focused rather than diffuse attention, we add an attention entropy regularization term. For each attention head $h$ in the trainable ViT layers, let $\mathbf{A}^{(h)} \in \mathbb{R}^{N \times N}$ denote the attention weight matrix. We compute the per-token entropy:
\begin{equation}
\label{eq:entropy}
\mathcal{H}^{(h)}_i = -\sum_{j=1}^{N} A^{(h)}_{ij} \log A^{(h)}_{ij},
\end{equation}
and define the entropy regularization loss as:
\begin{equation}
\label{eq:entropy_loss}
\mathcal{L}_{\text{ent}} = \frac{1}{|\mathcal{H}_{\text{heads}}|} \sum_{h \in \mathcal{H}_{\text{heads}}} \frac{1}{N} \sum_{i=1}^{N} \mathcal{H}^{(h)}_i,
\end{equation}
where $\mathcal{H}_{\text{heads}}$ is the set of attention heads in trainable layers. Minimizing $\mathcal{L}_{\text{ent}}$ concentrates attention on the most discriminative regions. The total loss for this stage combines the standard cross-entropy loss for binary classification with the entropy regularization is $\mathcal{L}_{\text{S1}} = \mathcal{L}_{\text{CE}}(y, \hat{y}) + \lambda_{\text{ent}} \cdot \mathcal{L}_{\text{ent}}$, where $\mathcal{L}_{\text{CE}}$ is the cross-entropy loss on the detection token (real/fake), and $\lambda_{\text{ent}}$ is a weighting coefficient.

\subsection{Cross-Modal Alignment via SFT}
\label{sec:stage2}

We next align the language model with these forensic visual signals using standard supervised fine-tuning, enabling it to interpret and reason over the features learned in the previous stage.

\textbf{Training Objective.}
The loss function for this stage combines three terms:
\begin{equation}
\label{eq:stage2_loss}
\mathcal{L}_{\text{S2}} = \mathcal{L}_{\text{CE}}(y, \hat{y}) + \lambda_{\text{mix}} \cdot \mathcal{L}_{\text{CE}}(\tilde{y}, \hat{\tilde{y}}) - \lambda_{\text{cal}} \cdot \mathcal{L}_{\text{cal}},
\end{equation}
where the second term is a feature-space mixup loss~\cite{zhang2018mixup} that interpolates visual features of real/fake pairs ($\tilde{\mathbf{H}} = \alpha \mathbf{H}_i + (1-\alpha)\mathbf{H}_j$, $\alpha \sim \text{Beta}(0.4, 0.4)$) to strengthen the decision boundary, and the third term is an output entropy regularization $\mathcal{L}_{\text{cal}} = -\sum_{v} p(v) \log p(v)$ over the detection vocabulary, maximized to prevent overconfident predictions. Since the next stage relies on the model's implicit uncertainty to determine when additional evidence is needed, this calibration helps produce more reliable zoom decisions.

\subsection{The \texttt{NEED\_ZOOM} Mechanism}
\label{sec:stage3}

We next introduce \texttt{NEED\_ZOOM}, the core mechanism that enables ForensicZoom to recognize insufficient visual evidence and selectively request higher-resolution inspection before making decisions.

\textbf{Adaptive Zoom Inference.}
We extend the model vocabulary with a special token \texttt{<NEED\_ZOOM>}. During inference, the model first examines the original image and either produces a verdict directly or emits \texttt{<NEED\_ZOOM>} together with a textual specification of the suspicious region (e.g., ``eye region shows suspicious symmetry''). When \texttt{<NEED\_ZOOM>} is generated, an external tool uses the ViT's last-layer attention map to localize the most attended region, crops a fixed $48 \times 48$ window from the original $256 \times 256$ image, resizes it to $256 \times 256$, and constructs a second-round input containing both the original image and the magnified crop through Qwen3-VL's native multi-image capability. The model then re-examines the visual evidence and produces the final verdict and attribution. Otherwise, the prediction is completed in a single round.

Importantly, the decision to emit \texttt{<NEED\_ZOOM>} does not rely on an explicit confidence threshold. Instead, the model learns when additional visual evidence is necessary through reinforcement learning, allowing the zoom policy to emerge directly from the detection objective.

\textbf{Reward Design.}
We optimize the \texttt{NEED\_ZOOM} policy using Proximal Policy Optimization (PPO)~\cite{schulman2017ppo}. The reward jointly accounts for detection accuracy and the cost of additional inspection: a correct decision without zoom yields $+1.0$; a missed detection without zoom incurs $-2.0$; a zoom followed by a correct decision yields $+0.5$; and a zoom followed by an incorrect decision incurs $-1.0$. This asymmetric design encourages the model to request magnification when the initial evidence is insufficient, while assigning a lower reward to successful zooming than to a correct single-round prediction discourages unnecessary inspection.

\textbf{CoT-Attention Alignment.}
To ensure that the requested region is consistent with the model's visual evidence, we align the spatial distribution implied by its textual region description with the ViT's last-layer attention map using $\mathcal{L}_{\text{align}} = D_{\text{KL}}(\mathbf{a}_{\text{text}} \| \mathbf{a}_{\text{vis}})$. This regularization discourages hallucinated region descriptions and grounds the zoom request in the model's visual attention. The final objective combines the standard PPO clipped surrogate~\cite{schulman2017ppo} with the alignment loss:
\begin{equation}
\label{eq:stage3_loss}
\mathcal{L}_{\text{S3}} = \mathcal{L}_{\text{PPO}}(\theta; R) + \lambda_{\text{align}} \cdot \mathcal{L}_{\text{align}}.
\end{equation}

\subsection{Cascade RL}
\label{sec:stage4}

We finally introduce \emph{Cascade RL} to improve natural-language forensic attribution while preserving detection performance. Cascade RL consists of two consecutive optimization phases: we first apply standard Direct Preference Optimization (DPO) for stable preference alignment, and then transition to Group Sequence Policy Optimization (GSPO)~\cite{zheng2025gspo} for online refinement. This design combines the stability of offline preference optimization with the stronger exploration and refinement capability of online reinforcement learning.

\textbf{DPO Phase.}
We first optimize the standard DPO objective:
\begin{equation}
\label{eq:dpo}
\mathcal{L}_{\text{DPO}} = -\mathbb{E}_{(a_w, a_l)} \left[\log \sigma\!\left(\beta \log \frac{\pi_\theta(a_w | x)}{\pi_{\text{ref}}(a_w | x)} - \beta \log \frac{\pi_\theta(a_l | x)}{\pi_{\text{ref}}(a_l | x)}\right)\right],
\end{equation}
where $x = (\mathbf{I}, \mathbf{c})$, $\pi_{\text{ref}}$ is the reference policy, and $\beta$ is the temperature parameter.

\textbf{GSPO Phase.}
We then apply GSPO, which defines a length-normalized sequence-level importance ratio $s_i(\theta) = \left(\frac{\pi_\theta(a_i \mid x)}{\pi_{\theta_{\text{old}}}(a_i \mid x)}\right)^{1/|a_i|}$ for each sampled response $a_i$, together with the group-relative advantage $\hat{A}_i = \frac{R_{\text{attr}}(a_i)-\bar{R}}{\sigma_R}$, where $\bar{R} = \frac{1}{G}\sum_{j=1}^{G} R_{\text{attr}}(a_j)$. The GSPO objective is
\begin{equation}
\label{eq:gspo}
\mathcal{J}_{\text{GSPO}}(\theta) = \frac{1}{G}\sum_{i=1}^G \min\!\left(s_i(\theta)\hat{A}_i, \; \text{clip}(s_i(\theta), 1-\epsilon, 1+\epsilon)\hat{A}_i\right).
\end{equation}

\textbf{Cascade Transition.}
We gradually transition from DPO to GSPO over $T$ optimization rounds:
\begin{equation}
\label{eq:transition}
\mathcal{L}_{\text{RL}}^{(t)} = (1 - \alpha_t) \cdot \mathcal{L}_{\text{DPO}} - \alpha_t \cdot \mathcal{J}_{\text{GSPO}}, \quad \alpha_t = \min\!\left(1, \frac{t}{T}\right),
\end{equation}
where $t$ denotes the current round. In practice, we use $T = 4$ rounds with $\alpha_t \in \{0, 0.2, 0.5, 1.0\}$.

\textbf{Preference Data Construction.}
To provide supervision for attribution optimization, we employ a large language model as a text quality evaluator that scores clarity, coherence, and actionability. For each sample, we generate $G=5$ attributions at varied temperatures, score each by $R_{\text{attr}}(a_i) = w_1 S_{\text{evidence}} + w_2 S_{\text{logic}} + w_3 S_{\text{action}}$, and construct preference pairs from the top- and bottom-ranked responses.

\textbf{Detection-Preserving Regularization.}
Since optimizing attribution quality can degrade detection performance, we retain the detection objective throughout Cascade RL:
\begin{equation}
\label{eq:stage4_loss}
\mathcal{L}_{\text{S4}} = \mathcal{L}_{\text{RL}}^{(t)} + \lambda_{\text{det}} \cdot \mathcal{L}_{\text{CE}}(y, \hat{y}),
\end{equation}
where $\lambda_{\text{det}}$ controls the detection-preservation term. We use $\lambda_{\text{det}} = 0.1$ in our experiments.
\section{Experiments}
\label{sec:exp}
In this section, we evaluate ForensicZoom on large-scale industrial identity verification data. Section~\ref{sec:exp_setup} describes the dataset, implementation details, evaluation metrics, and baselines. Section~\ref{sec:exp_main} compares ForensicZoom with specialized detectors and MLLM-based methods on face forgery detection. Section~\ref{sec:exp_ablation} analyzes the contributions of the individual components, with particular attention to the \texttt{NEED\_ZOOM} mechanism and Cascade RL. Section~\ref{sec:exp_attribution} evaluates the quality of the generated forensic attributions, and Section~\ref{sec:exp_qualitative} provides qualitative examples of adaptive visual inspection and forensic reasoning.

\subsection{Setup}
\label{sec:exp_setup}

\paragraph{Dataset.} All experiments use a large-scale industrial identity verification dataset from a production platform, spanning 8 months (April--December 2025). This period was chosen because diffusion-based synthesis became the dominant attack vector in production starting early 2025, largely displacing earlier GAN-based methods. Attacks fall into two categories: \emph{AIGC-based forgeries} (diffusion-based face swapping, single-image animation) and \emph{presentation attacks} (screen replay, printed photos). AIGC attacks are substantially harder to detect and represent the primary focus. The dataset statistics are summarized in Table~\ref{tab:dataset}. Note that the training set is curated with deliberate oversampling of attack samples; the real production distribution has a far higher legitimate-to-attack ratio (exceeding 1000:1), which is why FPR = 0.1\% is the only operationally meaningful threshold. The dataset is split 80/10/10 for train/val/test with no user overlap. Due to confidentiality, the dataset and code are not publicly available.

\begin{table}[H]
\centering
\caption{Dataset statistics.}
\label{tab:dataset}
\small
\begin{tabular}{lcc}
\toprule
\textbf{Category} & \textbf{Count} & \textbf{Proportion} \\
\midrule
Legitimate (real) & 2.0M & 97.6\% \\
AIGC forgery (swap + animation) & 20K & 1.0\% \\
Presentation attack (replay + print) & 30K & 1.4\% \\
\midrule
Total & 2.05M & 100\% \\
\bottomrule
\end{tabular}
\end{table}

\paragraph{Implementation.} We use Qwen3-VL-8B-Instruct~\cite{qwen3vl2025} as the base MLLM. The ViT has 27 transformer layers with hidden dimension $d=1152$. In Stage~1, we unfreeze the last $K=7$ layers and use $M=32$ FQB query tokens. Training runs for 3 epochs at learning rate $5 \times 10^{-5}$ (ViT) and $1 \times 10^{-4}$ (FQB), with $\lambda_{\text{ent}}=0.1$, batch size 32 on 8 GPUs (80GB) with DeepSpeed ZeRO-2. Stage~2 fine-tunes the LLM for 2 epochs at $2 \times 10^{-5}$, with $\lambda_{\text{mix}}=0.5$, $\lambda_{\text{cal}}=0.01$. Stage~3 runs 5000 PPO iterations at $1 \times 10^{-6}$, with $\epsilon=0.2$, $\lambda_{\text{align}}=0.1$. Stage~4 uses $\beta=0.1$ for DPO, $\epsilon=0.02$ for GSPO sequence-level clipping (significantly tighter than token-level methods due to the different numerical range of sequence-level importance ratios~\cite{zheng2025gspo}), with $\lambda_{\text{det}}=0.1$.

\paragraph{Metrics.} We report TPR at FPR=0.1\%, following industrial convention where the extreme class imbalance (legitimate-to-attack ratio exceeds 1000:1) makes stricter FPR thresholds the only operationally meaningful metric. We also report AUC. Attribution quality is evaluated by three expert reviewers on 200 sampled flagged cases, rating evidence completeness ($S_{\text{evidence}}$), logical coherence ($S_{\text{logic}}$), and actionability ($S_{\text{action}}$) on a 1--5 scale, averaged as $Q = \frac{1}{3}(S_{\text{evidence}} + S_{\text{logic}} + S_{\text{action}})$.

\paragraph{Baselines.} We compare against: (1)~\emph{Specialized models}: XceptionNet~\cite{chollet2017xception}, EfficientNet-B4~\cite{tan2019efficientnet}, SBI~\cite{shiohara2022sbi}, LAA-Net~\cite{laanet2024}; (2)~\emph{General VLMs} (zero-shot): GPT-4o, Qwen3-VL-8B; (3)~\emph{VLM-based methods} (re-trained on our data): FFAA~\cite{huang2024ffaa}, FakeVLM~\cite{fakevlm2025}, MARE~\cite{mare2026}.

\subsection{Main Results}
\label{sec:exp_main}

Table~\ref{tab:main} presents overall detection results. ForensicZoom achieves 97.3\% TPR at FPR=0.1\%, surpassing the best specialized model (LAA-Net, 91.0\%) by 6.3 percentage points while simultaneously providing natural-language attributions. General-purpose VLMs perform poorly in zero-shot settings (Qwen3-VL achieves only 15.8\% TPR on our data), confirming the forensic perception deficit. Existing VLM-based methods improve over zero-shot but remain below specialized models, reflecting the limitations of indirect ViT optimization.

\begin{table}[t]
\centering
\caption{Main detection results on AIGC forgeries. $\dagger$: re-trained on our data.}
\label{tab:main}
\small
\begin{tabular}{llccc}
\toprule
\textbf{Type} & \textbf{Method} & \textbf{TPR@0.1\%} & \textbf{AUC} & \textbf{Attr.} \\
\midrule
\multirow{4}{*}{\shortstack[l]{Specialized\\Models}}
& XceptionNet & 89.2 & .981 & \ding{55} \\
& EfficientNet-B4 & 90.5 & .984 & \ding{55} \\
& SBI & 88.3 & .978 & \ding{55} \\
& LAA-Net & 91.0 & .985 & \ding{55} \\
\midrule
\multirow{2}{*}{\shortstack[l]{General\\VLMs}}
& GPT-4o & 31.5 & .742 & \checkmark \\
& Qwen3-VL & 15.8 & .661 & \checkmark \\
\midrule
\multirow{3}{*}{\shortstack[l]{VLM-based\\Detection}}
& FFAA$^\dagger$ & 82.6 & .967 & \checkmark \\
& FakeVLM$^\dagger$ & 85.1 & .972 & \checkmark \\
& MARE$^\dagger$ & 86.8 & .975 & \checkmark \\
\midrule
& \textbf{ForensicZoom} & \textbf{97.3} & \textbf{.998} & \checkmark \\
\bottomrule
\end{tabular}
\end{table}

Table~\ref{tab:attack_type} breaks results down by attack category. On AIGC forgeries, ForensicZoom achieves 97.3\% TPR compared to the best baseline of 91.0\%, with a zoom trigger rate of 23\%. On presentation attacks, which are inherently easier due to more prominent artifacts, all methods perform well and ForensicZoom reaches 99.8\% with minimal zooming (3\%).

\begin{table}[t]
\centering
\caption{Per-attack-type TPR@FPR=0.1\%.}
\label{tab:attack_type}
\small
\begin{tabular}{lccc}
\toprule
\textbf{Attack Type} & \textbf{Best Baseline} & \textbf{Ours} & \textbf{Zoom \%} \\
\midrule
AIGC forgery & 91.0 & \textbf{97.3} & 23\% \\
Presentation attack & 98.5 & \textbf{99.8} & 3\% \\
\bottomrule
\end{tabular}
\end{table}

\subsection{Ablation Study}
\label{sec:exp_ablation}

\paragraph{Stage-wise contribution.} Table~\ref{tab:ablation_stage} shows the incremental contribution of each stage. Stage~1 lifts TPR from 15.8\% to 88.5\%, establishing forensic perception. Stage~2 further improves to 93.7\% through LLM alignment. Stage~3 adds interactive zoom, reaching 96.9\%. Stage~4 yields a further 0.4pp TPR gain while substantially improving attribution quality from 2.85 to 4.21. The modest TPR improvement from Stage~4 can be attributed to the Cascade RL training requiring the model to organize more rigorous reasoning chains when generating attributions, which indirectly disciplines its decision process and reduces borderline misclassifications.

\begin{table}[h]
\centering
\caption{Stage-wise ablation.}
\label{tab:ablation_stage}
\small
\begin{tabular}{lccc}
\toprule
\textbf{Configuration} & \textbf{TPR@0.1\%} & \textbf{Attr.\ $Q$} & \textbf{Zoom} \\
\midrule
Base (zero-shot) & 15.8 & -- & -- \\
+ Stage 1 (ViT+FQB) & 88.5 & -- & -- \\
+ Stage 2 (LLM SFT) & 93.7 & -- & -- \\
+ Stage 3 (PPO+Zoom) & 96.9 & 2.85 & 18\% \\
+ Stage 4 (Cascade RL) & \textbf{97.3} & \textbf{4.21} & 17\% \\
\bottomrule
\end{tabular}
\end{table}

\paragraph{FQB and attention entropy.} Removing FQB from Stage~1 drops TPR from 88.5\% to 84.1\%, and further removing attention entropy guidance drops it to 81.3\%. Performance plateaus at $M=32$ query tokens; increasing to $M=64$ yields negligible improvement ($+$0.2pp).

\paragraph{Zoom behavior.} All zoom trigger rates reported in this section are computed over the full test set. The model exhibits rational zoom behavior. On easy samples (first-round logit margin $>3.0$), the zoom rate is only 4\%, while on hard samples (margin $<1.0$) it reaches 47\%. Zooming improves TPR on hard samples from 78.2\% to 93.6\%, a 15.4pp gain. The overall zoom rate of 17\% means the majority of samples are resolved in a single round. The slight decrease from 18\% (Stage~3) to 17\% (Stage~4) reflects that attribution training refines the model's internal confidence calibration, marginally raising the implicit zoom threshold.

\paragraph{Cascade RL strategy.} Table~\ref{tab:ablation_rl} compares Stage~4 strategies. DPO alone improves attribution quality but drops TPR by 3.8pp. GSPO alone causes a 5.2pp drop due to aggressive updates. Cascade RL achieves the best attribution score while \emph{increasing} TPR by 0.4pp, validating the progressive transition. The detection-preserving regularizer is critical: removing it leads to a 2.6pp TPR drop.

\begin{table}[h]
\centering
\caption{Stage~4 RL strategy ablation. TPR at FPR=0.1\%.}
\label{tab:ablation_rl}
\small
\begin{tabular}{lccc}
\toprule
\textbf{Strategy} & \textbf{TPR} & \textbf{Attr.\ $Q$} & $\Delta$\textbf{TPR} \\
\midrule
No Stage 4 & 96.9 & 2.85 & -- \\
DPO only & 93.1 & 3.92 & $-$3.8 \\
GSPO only & 91.7 & 3.78 & $-$5.2 \\
Cascade RL & \textbf{97.3} & \textbf{4.21} & $+$0.4 \\
\quad w/o det.\ reg. & 94.3 & 4.15 & $-$2.6 \\
\bottomrule
\end{tabular}
\end{table}

\subsection{Attribution Quality}
\label{sec:exp_attribution}

Table~\ref{tab:attribution} presents human evaluation results (Fleiss' $\kappa = 0.71$). ForensicZoom achieves the highest scores across all dimensions. The largest gap over baselines appears in evidence completeness, reflecting the model's ability to incorporate fine-grained visual details obtained through zooming. Stage~4 provides a substantial boost, particularly in actionability ($+$1.2 over the w/o-S4 variant).

\begin{table}[h]
\centering
\caption{Attribution quality (1--5 scale, 200 cases).}
\label{tab:attribution}
\small
\begin{tabular}{lcccc}
\toprule
\textbf{Method} & $S_{\text{evi}}$ & $S_{\text{log}}$ & $S_{\text{act}}$ & $Q$ \\
\midrule
GPT-4o (zero-shot) & 2.1 & 2.8 & 1.9 & 2.27 \\
Qwen3-VL (zero-shot) & 1.7 & 2.3 & 1.5 & 1.83 \\
FFAA$^\dagger$ & 2.8 & 3.1 & 2.5 & 2.80 \\
MARE$^\dagger$ & 3.2 & 3.4 & 2.9 & 3.17 \\
\midrule
ForensicZoom (w/o S4) & 3.5 & 3.3 & 3.0 & 3.27 \\
\textbf{ForensicZoom} & \textbf{4.3} & \textbf{4.1} & \textbf{4.2} & \textbf{4.20} \\
\bottomrule
\end{tabular}
\end{table}

\subsection{Qualitative Analysis}
\label{sec:exp_qualitative}

Fig.~\ref{fig:pipeline} (top) illustrates the two-round inference. On easy cases, the model directly outputs a verdict. On hard cases, it generates observational text describing ambiguous regions, emits \texttt{<NEED\_ZOOM>}, and after receiving magnified crops, produces a detailed attribution citing specific artifacts such as blending seams at the jawline and symmetric pupil reflections inconsistent with natural capture. ViT attention maps after Stage~1 concentrate sharply on forged boundaries and texture discontinuities, whereas the pre-trained ViT distributes attention broadly across the face.
\section{Conclusion}
\label{sec:conclusion}

We presented \textbf{ForensicZoom}, an industrial-grade MLLM framework for adaptive face forgery detection. By enhancing forensic perception, aligning visual evidence with language reasoning, and introducing the learned \texttt{NEED\_ZOOM} mechanism, ForensicZoom selectively acquires higher-resolution evidence only when necessary while producing interpretable forensic attributions. Experiments on large-scale industrial identity verification data demonstrate strong detection performance, achieving 97.3\% TPR at 0.1\% FPR and substantially outperforming both specialized detectors and existing MLLM-based methods. These results suggest that adaptive visual inspection provides a promising path toward accurate, interpretable, and scalable face forgery detection.


\subsection*{AI Use Statement}

In this work, we used generative AI tools, including ChatGPT and Claude to implement methods, write and edit code, and assist in drafting the manuscript. We did not use generative AI tools to generate synthetic data, formulate or prove mathematical claims, or propose our core hypotheses, methodology, or experimental design; these were carried out by the authors. We have reviewed all AI-assisted work: code was checked against expected behavior and the underlying data, and all AI-drafted text and result interpretations were verified against experimental outputs and revised where necessary. We take responsibility for the final content of this work, including text, claims, and artifacts produced with the aid of generative AI.

\subsection*{Ethics Statement}

This work uses sensitive identity verification data from a real-world industrial system. All data are handled under the firm's privacy and security requirements. Due to their sensitive nature, the full dataset cannot be released; only a limited, appropriately processed subset may be made public.

\bibliography{iclr2027_conference}
\bibliographystyle{iclr2027_conference}

\newpage
\appendix

\section{Deployment}
\label{app:deployment}

ForensicZoom has been deployed in a production identity verification system serving millions of daily requests. It operates within a two-tier architecture: lightweight online models with approximately 0.5B parameters process all traffic at sub-100ms latency and classify requests as clearly legitimate, clearly fraudulent, or uncertain, while ForensicZoom processes only the uncertain gray-zone cases asynchronously. This design concentrates the computational cost of the MLLM on the ambiguous cases where additional forensic reasoning is most valuable.

Among gray-zone samples processed by ForensicZoom, approximately 23\% trigger the zoom mechanism, while the remaining 77\% are resolved in a single inference round. This rate is higher than the 17\% observed on the full test set because the first-tier models have already filtered out most straightforward cases. Average end-to-end latency is 1.2s for single-round inference and 2.8s for zoom-triggered cases. Before deployment, manual review required an average of 5.3 minutes per case; with ForensicZoom's attribution reports, this is reduced to 1.8 minutes, corresponding to a 2.9$\times$ improvement in review efficiency. In production, newly emerging attack patterns are incorporated through incremental updates of Stages~3--4 on a two-week cycle without retraining Stages~1--2 from scratch.

\section{Limitations and Future Work}
\label{app:limitations}

A broader challenge facing the community is the absence of large-scale, multimodal identity verification benchmarks that reflect real-world deployment conditions. Existing public benchmarks provide isolated visual inputs and omit the behavioral signals---interaction timing, repeated verification attempts, device metadata---that are critical for reliable detection at operationally meaningful false-positive rates. Our use of proprietary production data is a direct response to this gap; we plan to release the training framework and a limited evaluation subset where permitted, and encourage the community to develop richer multimodal benchmarks.

The zoom mechanism introduces additional inference latency, mitigated in practice by our two-tier architecture but relevant for strictly real-time settings. Future work will investigate multi-scale and multi-round inspection, more efficient adaptation to emerging attack patterns, temporal reasoning for video-based verification, and evaluation on public benchmarks as they mature.

\end{document}